\PassOptionsToPackage{table, svgnames, dvipsnames}{xcolor}
\documentclass[final]{anthology-ch} 

\usepackage{booktabs}
\usepackage{graphicx}
\usepackage{arydshln}
\usepackage{enumitem}
\usepackage{algorithm,amsmath,algpseudocode}
\usepackage{bbding}
\usepackage{xcolor}

\usepackage{makecell, cellspace, caption}

\newcommand{\greencheck}{\textcolor{green!60!black}{\checkmark}}
\newcommand{\redtimes}{\textcolor{red!75!black}{\boldmath$\times$}}

\title{Continuous sentiment scores for literary and multilingual contexts}

\author[1, *]{Laurits Lyngbaek}[  orcid=0009-0008-7995-3459]

\author[1, *]{Pascale Feldkamp}[  orcid=0000-0002-2434-4268]

\author[1]{Yuri Bizzoni}[  orcid=0000-0002-6981-7903]

\author[1]{Kristoffer L. Nielbo}[  orcid=0000-0002-5116-5070]

\author[1]{Kenneth Enevoldsen}[  orcid=0000-0001-8733-0966]

\affiliation{1}{Center for Humanities Computing, Aarhus University, Aarhus, Denmark}
\affiliation{*}{Joint First Authorship}

\pubyear{2025}
\pubvolume{1}
\pagestart{1}
\pageend{1}
\conferencename{Proceedings of Conference XXX}
\conferenceeditors{Editor1 Editor2}
\doi{00000/00000}  

\begin{document}

\maketitle


\begin{abstract}
Sentiment Analysis is widely used to quantify sentiment in text, but its application to literary texts poses unique challenges due to figurative language, stylistic ambiguity, as well as sentiment evocation strategies. Traditional dictionary-based tools tend to underperform, especially for low-resource languages, and transformer models, while promising, output coarse categorical labels that limit fine-grained analysis. We introduce a novel continuous sentiment scoring method based on concept vector projection, trained on multilingual literary data, which captures nuanced sentiment expressions across genres, languages, and historical periods. Our approach outperforms existing tools on English and Danish texts, producing sentiment scores which distribution matches human ratings, improving sentiment arc modeling and analysis in literature.
\end{abstract}

\section{Introduction \& Related Works}

Sentiment analysis quantifies sentiment in text and is widespread across domains, from product reviews analysis to social media monitoring \autocite{tsao_asymmetric_2018, bollen_twitter_2011}. Computational literary studies have employed sentiment analysis to model narrative dynamics, particularly sentiment arcs, across novels \autocite{jockers_novel_2014, reagan_emotional_2016, cellier_prediction_2016, bizzoni_sentimental_2023}. This requires continuous sentiment scores, mapping sentiment along a spectrum rather than using categorical labels like positive/negative.

Despite the growing use of continuous sentiment scoring in literary studies, the validity of current tools in capturing literary sentiment expression remains underexplored. Popular tools such as \textit{Syuzhet} have faced severe criticism for oversimplification or poor generalizability \autocite{swafford2015problems} -- issues that point to broader limitations in applying off-the-shelf sentiment tools to literary texts. Indeed, the literary domain poses distinct challenges: figurative language, multiple narrative layers, and stylistic ambiguity all complicate sentiment detection \autocite{booth_rhetoric_1983, bizzoni_feldkamp_sea_2024}.

More recent transformer-based models appear better equipped to handle the complexity of literary language \autocite{schmidt_using_2021}, and techniques exist to transform categorical model outputs into continuous scores \autocite{bizzoni_comparing_2023}. This method has proven more effective than tailored dictionary-based tools, particularly in low-resource language settings and across languages \autocite{feldkamp_comparing_2024}.
However, empirical benchmarks comparing model predictions to human judgments remain limited in languages other than English.

\vspace{.5em}

We identify three main issues where current methods see a noticeable performance drop:
\begin{enumerate}[noitemsep,topsep=0pt,wide=0pt,label=\textbf{\arabic*})]
    \item \textbf{Cross-lingual} performance drops. Most Sentiment Analysis tools tackle high-resource languages, and their transfer to low-resource ones like Danish is non-trivial. Although Danish has several dictionary-based tools (i.a., Asent \cite{Enevoldsen_Asent_Fast_flexible_2022}, Sentida \cite{lauridsen_sentida_2019}), these have seen little use on historical literature and struggle with complex literary forms. Comparing tools for Danish literary sentiment analysis, \textcite{feldkamp_comparing_2024} found that multilingual transformer models outperfromed dictionaries -- likely because they leverage contextual attention. While multilingual transformers, such as mBERT and XLM-R \autocite{Devlin_2019}, show promise for cross-lingual sentiment analysis in literature \autocite{feldkamp_comparing_2024}, cultural and linguistic biases inherited from English pretraining remain a concern \autocite{de_bruyne_2022_language, xu_survey_2022}.

    \item \textbf{Cross-domain} performance often drops, especially when applying tools trained on social media to literature, where sentiment is expressed in a distinct and complex manner \autocite{bizzoni_feldkamp_sea_2024, feldkamp2024sentiment, vishnubhotla-etal-2024-emotion}. Literary language tends to be more \textit{omissive and implicit}, relying less on charged vocabulary and more on \textit{concrete} descriptions of objects and situations to evoke affect -- a domain-specific mode of sentiment expression that models fail to capture \autocite{feldkamp2024sentiment}.
    This domain-specificity varies across domains: when using a model fine-tuned on Twitter posts, poetry shows the weakest correlation with human ratings, prose falls in the middle, and Facebook posts show the strongest correlation \autocite{feldkamp2024sentiment}.

    \item \textbf{Historical data}, marked by diachronic language change, reduces model performance. While fine-tuned multilingual transformers show promise \autocite{allaith_sentiment_2023, schmidt_evaluation_2018, feldkamp_comparing_2024}, challenges remain. Lexical drift -- including semantic shift, word loss (e.g., \textit{thou}, \textit{peradventure}), changing frequencies, and temporal polarity shifts -- limits sentiment inference if models rely on priors from modern corpora.\footnote{Diachronic sentiment analysis is challenging for traditional machine learning approaches as words’ meaning and polarity change in a continuous way, while most models require steady ground truths for training, creating artificial ``museums'' of words’ sentiment scores in a given historical period.}  For temporal polarity shifts, even short-term changes can lower model performance \cite{lukes_sentiment_2018}. 
    
\end{enumerate}

\vspace{.5em}

A major drawback of recent transformer-based approaches is that, while they outperform dictionary-based tools on historical and literary data \autocite{feldkamp_comparing_2024}, they tend to perform trinary classifications (positive, neutral, negative), limiting their usefulness for fine-grained sentiment analysis.
Although model confidence scores can be repurposed for continuous output -- with medium to strong correlation to human ratings \autocite{feldkamp_comparing_2024} -- the resulting distributions still cluster around the original three categories, producing what is effectively a pseudo-trinary distribution. This poses a problem for literary analysis tasks, not least sentiment arc modelling, where detrending methods to smoothen out the signal necessitate continuous scores. 
When sentiment scores behave in extreme ways -- as they will with pseudo-trinary distributions -- smoothing will tend to collapse variation toward the neutral midpoint, removing meaningful information.

In this paper, we introduce a method to create continuous-scale sentiment scores that are more closely aligned with the distribution of human scores, while also mitigating language-, domain-, and historical data issues by basing the method on the language and domain of the use case.

We test this approach on English and Danish literary texts, comparing it to existing transformer-based models and popular dictionary-based tools, across both fiction and nonfiction genres. The benchmark includes both historical literary genres (e.g., 19\textsuperscript{th}-century hymns) and contemporary texts (e.g., blogs), enabling us to evaluate model performance in settings that better reflect the needs of researchers working with multilingual or diachronic literary corpora.
We pursue three aims: (1) to assess model performance on contemporary literary and non-literary texts; (2) to compare performance across literary genres; and (3) to evaluate models on historical and multilingual literary data. We begin by testing our approach on \textit{Fiction4} — a recent annotated fiction corpus that spans four literary genres, two languages (English and Danish), in the period 1798 to 1965. We then validate our approach further on \textit{EmoBank}, a standard sentiment analysis dataset that includes contemporary genres and a small set of fiction, to gauge the generalizability of our approach and to control for overfitting to literary data.

\section{Methods}
\subsection{Data}

\begin{table}[htbp]
    \centering
    \small
    \begin{tabular}{lcrrrrr}
    \toprule
    \textbf{Dataset} &  \textbf{Period} & \textbf{N annotations} & \textbf{N words} & \textbf{$\bar{x}$ words/sentence} & \textbf{N annotators} \\ 
    \midrule
       \textit{$\downarrow$ EmoBank} & 1990-2008 &  8,870 & 143,499 & 16.18 & 10  \\ \midrule
        \hspace{1.5em}Letters & & 1,413 & 21,639 & 15.31  & 10  \\
        \hspace{1.5em}Blog & & 1,336 &  20,874 & 15.62 & 10  \\
        \hspace{1.5em}Newspaper & & 1,314 & 25,992 & 19.78 & 10  \\
        \hspace{1.5em}Essays & & 1,135 & 26,349 & 23.21 & 10  \\
        \hspace{1.5em}Fiction & & 2,753 & 31,491 & 11.44 & 10  \\
        \hspace{1.5em}Travel-guides & & 919 & 17,154 & 18.67 & 10 \\
        \midrule
        \midrule
       \textit{$\downarrow$ Fiction4} & 1798-1965 & 6,300 & 73,250 & 11.6 & $>$2  \\ \midrule
        \hspace{1.5em}Hymns & 1798-1873 & 2,026 & 12,798 & 6.3 & 2 \\
        \hspace{1.5em}Fairy tales & 1837-1847 & 772 & 18,597 & 24.1 & 3\\
        \hspace{1.5em}Prose & 1952 & 1,923 & 30,279 & 15.7 & 2  \\
        \hspace{1.5em}Poetry & 1965 & 1,579 & 11,576 & 7.3 & 3  \\        
        \bottomrule
    \end{tabular}
        \caption{Datasets with valence annotation. Valence was annotated on a sentence basis, so `N annotations' indicates the number of sentences. The total number of sentences considered is $n=15,170$. `N annotators' indicates the number of annotators reported per sentence.}
    \label{tab:dataset_stats}
\end{table}

\paragraph{Emobank} is a text corpus manually annotated for sentiment according to the psychological Valence-Arousal-Dominance scheme. It was compiled at \textit{JULIE Lab}, Jena University \autocite{buechel-hahn-2017-emobank},\footnote{\url{https://github.com/JULIELab/EmoBank/}}, containing sentences from the MASC dataset, which is diverse both in terms of overall composition with diverse domains, and topically within categories.\footnote{On some \textit{EmoBank} categories: \textit{Essays} includes eight texts, i.a., ``A Brief History of Steel in Northeastern Ohio'. \textit{Fiction} comprises six prose pieces across genres, i.a., Richard Harding's ``A Wasted Day'' and the SciFi story ``Captured Moments''. \textit{Newspapers} contain reports (e.g., ``A.L. Williams Corp. was merged into Primerica Corp.'') and longer reportages. \textit{Travel Guides} are written in prose, including both place histories (e.g., ``A Brief History of Jerusalem'') and reflective pieces (e.g., ``Dublin and the Dubliners''). See the full MASC corpus at: \url{https://anc.org/data/masc/corpus/browse-masc-data/.}} It includes six categories: Letters, Blog, Newspaper, Essays, Fiction, and Travel guides.\footnote{We excluded the `Sem-Eval' category as it was internally diverse.}
Inter Rater Reliability for the whole dataset is: Krippendorff's $\alpha=0.34$.\footnote{Since \textit{EmoBank} lacks unique annotator IDs, we cannot correlate individual annotators' scores. Instead, Krippendorff’s $\alpha$ measures agreement across ratings per item. IRR per subset is shown in \autoref{tab:EmoBank_results}.}
We use the mean sentence-based valence scores overall and per category to compare model performance.

\paragraph{Fiction4} is a dataset of literary texts, spanning literary texts across four genres and two languages (English and Danish) in the 19\textsuperscript{th} and 20\textsuperscript{th} century.\footnote{ \url{https://huggingface.co/datasets/chcaa/fiction4sentiment}, for details, see \autocite{feldkamp2024sentiment}}, compiled at the \textit{Center for Humanities Computing}, Aarhus University. 
The corpus consists of three main authors, Sylvia Plath for poetry, Ernest Hemingway for prose, and H.C. Andersen for fairytales. Hymns were collected from Danish official church hymnbooks published between 1798 and 1873. All sentences in the corpus were annotated for by at least two annotators \autocite{feldkamp2024sentiment}. Inter Rater Reliability for the whole dataset is: Spearman's $\rho=0.63$ and Krippendorff's $\alpha=0.67$.\footnote{Humans rarely reach an agreement higher than 80\% ($\alpha$>0.80) for categorical tagging (positive/neutral/negative) on \textit{nonliterary texts} \autocite{wilson_recognizing_2005} -- and have lower IRR for continuous scale annotation \autocite{batanovic_versatile_2020} -- especially of literary texts \autocite{rebora_comparing_2023}.} We use the mean sentence-based valence score overall, per language set, and per genre to compare model performance.

\subsection{Comparison models}


\subsubsection{Dictionary-based}
\label{sec:dictionary_details}
Because of their popularity and wide usage in literary studies \autocite{allaith_sentiment_2023, bizzoni_nobels, bizzoni_comparing_2023}, as a baseline, we tested the dictionary-based tools \texttt{VADER} \autocite{hutto_vader_2014} and \texttt{Syuzhet} \autocite{syuzhet}. They assign sentiment scores (from negative to positive) by word-score matching and specific rules. \texttt{Syuzhet} was developed explicitly for literary texts.\footnote{The \texttt{Syuzhet} lexicon was developed in the \textit{Nebraska Literary Lab} under the direction of Matthew L. Jockers.}
When using these tools, we translated Danish sentences into English as they do not perform well on the original Danish.\footnote{Using \texttt{googletrans}: \url{https://pypi.org/project/googletrans/}. Humans did not review translations.} As such, dictionaries represent a rough baseline.

\subsubsection{Transformer-based}
\label{sec:transformer_scores_methods}
To test transformer-based methods, we chose two multilingual models. When testing models on Danish texts, we added three models fine-tuned for Danish. These were all tested across \textit{EmoBank} categories, as well as \textit{Fiction4} genres and languages. We list all models in \autoref{appdx:A}, \autoref{tab:modelnames}.\footnote{Code for comparing (HuggingFace-stored) sentiment models (with transformed outputs) on the \textit{Fiction4} or \textit{EmoBank} is at: \url{https://github.com/centre-for-humanities-computing/literary_sentiment_benchmarking}.} One of the multilingual models -- \texttt{twitter-xlm} -- showed the best performance on \textit{Fiction4} in \textcite{feldkamp_comparing_2024}. Danish models were picked based on their performance in a recent benchmark \autocite{allaith-etal-2023-sentiment}, and -- in the case of \texttt{MeMo-BERT-Sa} -- for being developed for 19\textsuperscript{th}-century novels \autocite{al-laith_development_2024}.

\textit{Conversion of model output:}
We convert Transformers' standard three-ways outputs (positive, neutral, negative) to continuous values using their confidence scores \footnote{The score output by finetuned models (e.g., ``positive'', 0.66) is a softmax-normalized class probability -- a pseudo confidence score -- reflecting how strongly a model prefers one label over another. It comes from the linear classification head atop models.} as a proxy for intensity (e.g., \textit{positive}, 0.67 → +0.67; \textit{negative} → –0.67; \textit{neutral} → 0). Mapping a model’s confidence values to a continuous scale often outperforms dictionary-based tools for literary sentiment \cite{bizzoni_comparing_2023, feldkamp_comparing_2024}.

\small
\[
\text{intensity} =
\begin{cases}
+p, & \text{if positive},\\
0,  & \text{if neutral},\\
-p, & \text{if negative}.
\end{cases}
\]
\normalsize

We tested transformer-based models on the original Danish and English, as well as on the Danish sentences translated to English (see \autoref{tab:sentiment_results_language}), since one study found that some models work better on google-translated sentences \cite{feldkamp_comparing_2024}, perhaps as the translation acts to standardize historic forms.

\subsection{Our approach}
\begin{figure}[ht]
    \centering
    \includegraphics[width=1\linewidth]{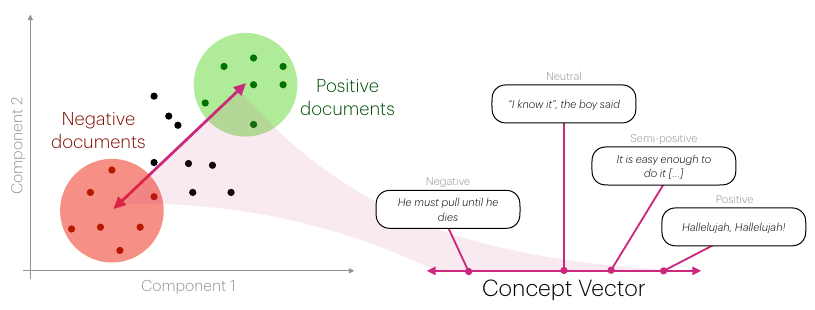}
    \caption{An overview of how a concept vector for sentiment is constructed and what information it contains. A circle represents an embedded document.}
    \label{fig:concept-vector-intro}
\end{figure}

It has been claimed that concepts -- such as a sentiment -- are approximately represented in a linear fashion within embedding space, which is denoted by the linear representation hypothesis \autocite{park_linear_2024}. The hypothesis states that concepts are encoded as a direction in the embedding space and that the further you move in a given direction, the stronger the concept is represented (see \autoref{fig:concept-vector-intro}). These linear representations of semantic information have been found in both encoding and decoding models, at varying levels of abstraction \autocite{wehner_taxonomy_2025, vu-parker-2016-k, li-etal-2021-implicit, zhao2024beyond}. Suppose we have access to the direction that encodes sentiment. In that case, we can project any embedded sentence onto the concept vector and gauge the sentiment of any given sentence, as seen in \autoref{fig:concept-vector-intro}. 

\subsubsection{Concept Vector Projection}
We propose an algorithm that constructs a concept vector in a given embedding space using positive and negative exemplary sentences that represent the opposing extremes of the concept. Using a pre-trained sentence embedding model $\mathbf{M}$, the algorithm embeds a labeled set of sentences $\mathbf{S}$. It assumes that a concept -- here sentiment -- is represented linearly in the embedding space. To define the concept vector, the algorithm computes the mean embedding of both the positive and negative sentiment examples, then calculates the vector pointing from the negative to the positive mean. This relies on the assumption that when averaging multiple sentences, all non-sentiment information will disappear as Gaussian noise with a mean of zero, leaving behind only the information encoding sentiment \autocite{kim2018interpretability, zhao2024beyond}.

\begin{figure}[htbp]
    \centering
    \includegraphics[width=.8\linewidth]{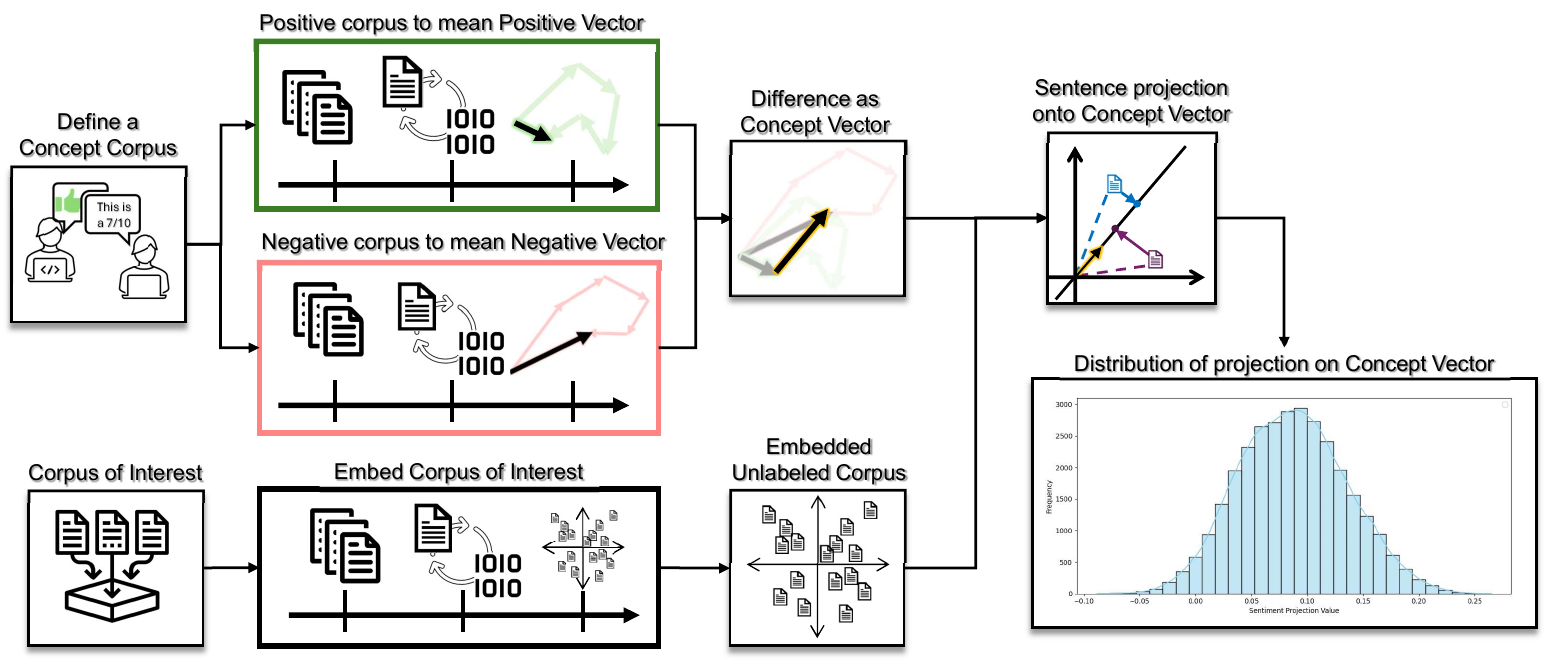}
    \caption{A visualization of how the Concept Vector Projection is constructed. It shows how to use a labeled sentiment corpus to predict sentiments of an unlabeled corpus of interest. The vectors shown are reduced to a two-dimensional Euclidean space for visualization, but normally reside in a high-dimensional space.}
    \label{fig:projection_workflow}
\end{figure}

The resulting vector then theoretically encodes sentiment direction. New sentences can be assessed for their relation to the sentiment by projecting their embeddings onto this vector: the farther along the direction the projection lies, the stronger their positive relation is. Defining the concept vector as a unit vector, the projection of a given embedding $ \mathbf{e}_i$ onto the unit concept vector $\hat{\mathbf{v}}$ is given by the dot product: $\mathbf{e}_i \cdot \hat{\mathbf{v}}$. This projects the sentence embedding to the subspace spanned by the Concept Vector. The high-dimensional embedding has thereby been reduced to a one-dimensional sentiment score, as seen in figure \autoref{fig:projection_workflow}. Defining a concept vector requires only a set of positive and negative example sentences. This suffices to predict the sentiment of any subsequent sentence, whether labeled or unlabeled. The Concept Vector Projection (CVP) algorithm formally described in \autoref{sec:algorithm}. The implementation of this method is available at \url{https://github.com/centre-for-humanities-computing/embedding-projection}. 

\subsection{Models}
The implementation of Concept Vector Projection used to classify sentiment in this paper is based on the language model \texttt{paraphrase-multilingual-mpnet-base-V2}\footnote{\url{https://huggingface.co/sentence-transformers/paraphrase-multilingual-mpnet-base-v2}} \autocite{reimers-2019-sentence-bert}. This is a 278M parameter model, based on a mean-pooled BERT architecture, optimized for sentence similarity by using Siamese and Triplet networks. This model was chosen because of its multilingual capabilities and excellent size-to-performance ratio. Investigations during model selection indicate that a larger model may increase model correlation with human ratings in exchange for compute budget.

Our Concept Vector was defined using a training dataset of sentences with positive and negative sentiments from the \textit{Fiction4} dataset. Since the sentences were originally rated on a numerical scale (1-9), they were translated to positive/negative ratings for the algorithm. We converted the mean ratings into ordinal labels through preset thresholds. That is, for the \textit{Fiction4} ratings, we define:

\small
$$\text{label} =
\begin{cases}
\text{positive}^+ & \text{if rating} \geq 7 \\
\text{neutral}^\varnothing & \text{if } 7 > \text{rating} > 3 \\
\text{negative}^- & \text{if rating} \leq 3
\end{cases}
]$$
\normalsize

All the neutral sentences and 60\% of the positive and negative sentences were in the \textit{Fiction4} testing set. The remaining 40\% were in a Concept Corpus of 204 positive and 168 negative sentences used to define the model's concept vector.

\section{Results}

\subsection{Continuous scoring}

\begin{figure}[h!]
    \centering
    \includegraphics[width=1\linewidth]{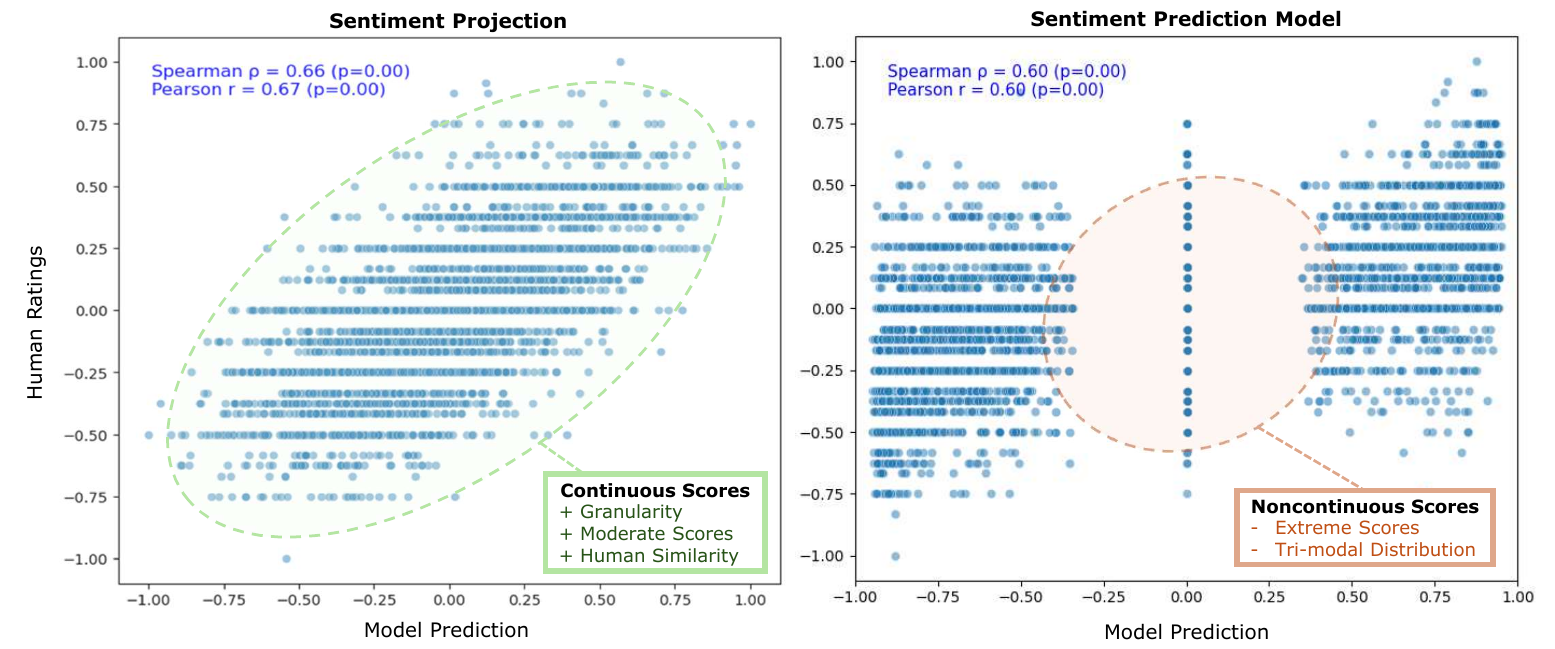}
    \caption{Scatterplot of Sentiment Predictions for respectively \texttt{Sentiment Projection} and \texttt{xlm-roberta}. While the \texttt{xlm-roberta} model, in theory, can predict a continuous space of sentiments when transforming it with confidence scores, inspection shows that certain ranges of the sentiments spectrum are not used. While both models achieve high correlations, it appears that \texttt{xlm-roberta} achieves this by matching human tendencies to predict neutral.}
    \label{fig:scatterplots_corr}
\end{figure}

A key benefit of the Sentiment Projection model is its ability -- like dictionary tools -- to produce genuinely continuous predictions. 
In contrast, Transformer-based token-classification models such as \texttt{xlm-roberta}, which can be coerced to output continuous scores (see \autoref{sec:transformer_scores_methods}), in practice exhibit a ``pseudo-trinary'' behavior: their predictions cluster heavily at zero and at the two polar extremes. 
This behavior is visible both in the scatterplots of predicted vs true sentiments (\autoref{fig:scatterplots_corr}) and in the histograms of model outputs (Appendix A, \autoref{fig:Prediction_Histograms}). When looking at the \textit{EmoBank} results (Appendix A, \autoref{fig:EmoBank_Scatter}), the discretized output of \texttt{xlm-roberta} appears even more sharply tri-modal than the human scores, which average ten annotators.

\subsection{Performance on literary data across genres}

\autoref{tab:fiction4_corr_per_genre} compares our model’s predictions to the human gold-standard ratings for the \textit{Fiction4} dataset's 4 genres. 

\begin{table}[h!]
    \centering
    \footnotesize
    \begin{tabular}{ll|c|c|cccc}
    \toprule
    \textbf{Type} & \textbf{Model} & \textbf{Scalar} & \textbf{Overall} & \textbf{Hymns} & \textbf{Fairy tales} & \textbf{Prose} & \textbf{Poetry} \\
    \midrule
    \midrule
    Year & & & & 1798--1873 & 1837--1847 & 1952 & 1965 \\
    \midrule
    \midrule
    Human $\rightarrow$ & IRR $\rho$          & \greencheck & 0.63 & 0.73 & 0.68 & 0.62 & 0.59 \\
                        & IRR $\alpha$       & \greencheck & 0.67 & 0.72 & 0.68 & 0.61 & 0.58 \\
    \midrule
    \midrule
    $\downarrow$ Dictionary   & \texttt{vader}     & \greencheck & 0.49 & 0.52 & 0.50 & 0.43 & 0.46 \\
                              & \texttt{syuzhet}   & \greencheck & 0.50 & 0.54 & 0.48 & 0.45 & 0.49 \\
    \midrule
    $\downarrow$ Multiling.    & \texttt{twitter-xlm}         & \redtimes  & 0.55 & 0.50 & 0.52 & 0.57 & \underline{0.58} \\
                              & \texttt{xlm-roberta}        & \redtimes  & \underline{0.60} & 0.59 & 0.62 & \underline{0.61} & \underline{0.57} \\
    \rowcolor{Gainsboro!35}
                              & \texttt{Sentiment Projection} & \greencheck & \textbf{0.66} & \textbf{0.69} & \underline{0.66} & \textbf{0.62} & \textbf{0.70} \\
    \midrule
    $\downarrow$ Danish        & \texttt{danish-sentiment}    & \redtimes  & 0.54 & 0.49 & 0.48 & 0.57 & 0.57 \\
                              & \texttt{da-sentiment-base}   & \redtimes  & 0.23 & 0.44 & 0.47 & 0.08 & 0.08 \\
                              & \texttt{MeMo-BERT-SA}        & \redtimes  & 0.47 & \underline{0.63} & \textbf{0.72} & 0.26 & 0.16 \\
    \bottomrule
    \end{tabular}
        \caption{Spearman correlations in the \textit{Fiction4} corpus \textit{across genres}. \underline{From top to bottom}: Publication years; then Inter Rater Reliability (human scores) per genre (Spearman's $\rho$ and Krippendorff's $\alpha$); then correlation between the human gold standard and models (Spearman's $\rho$). For VADER and Syuzhet scores, texts were automatically translated into English.}
    \label{tab:fiction4_corr_per_genre}
\end{table}

We evaluated all models on the full multilingual \textit{Fiction4} corpus. For the dictionary-based tools (\texttt{VADER} and \texttt{Syuzhet}), originally Danish texts were translated into English (see \autoref{sec:dictionary_details}). Danish-specific models generally under-perform on genres that are (originally) in English (\textit{Prose, Poetry}), which drags down their overall correlation scores. An outlier is \texttt{danish-sentiment}, which delivers relatively consistent results across both languages; however, it still falls short of \texttt{MeMo-BERT-SA} on the original Danish texts -- most notably in the Fairy Tales genre.

Most Danish transformer-based models perform on par with (or worse than) dictionary-based models applied to English translations of the original Danish texts (e.g., Fairy tales \& Hymns). Sentiment Projection, in contrast, achieves the highest correlation on every genre except Fairy tales -- where \texttt{MeMo-BERT-SA} performed best, which aligns with its fine-tuning on Danish literary prose from H.C. Andersen’s period. It performs especially well on Poetry, where other models struggle.

The genres that achieved the highest human IRR -- like hymns, at IRR $\rho=0.77$ -- did not reflect in better results for most models. The second-best performing model, \texttt{xlm-roberta}, for example, placed second-to-last on hymns. Instead, Sentiment Projection meets or exceeds Inter Rater correlation ($\rho$) for all genres.

\subsection{Performance on literary data across time and languages}

Results for the multilingual performance assessment are presented in \autoref{tab:sentiment_results_language}.

\begin{table*}[ht]
\centering
\footnotesize
\begin{tabular}{ll|c|ccc|c}
\toprule
\textbf{Type} & \textbf{Model} & \textbf{Scalar} & \textbf{Multiling.} & \textbf{Danish set} & \textbf{ English set} & \textbf{Translated}\\
& & & [Da + En] & [Da] & [En] & [Da $\rightarrow$ En] \\
\midrule
Human $\rightarrow$ & IRR $\rho$ & \greencheck & 0.63 & 0.68 & 0.58 & - \\
& IRR $\alpha$ & \greencheck & 0.67 & 0.71 & 0.60 & - \\
\midrule
\midrule
$\downarrow$ Dictionary & \texttt{vader} & \greencheck & - & - & 0.45 & 0.51 \\
& \texttt{syuzhet} & \greencheck & - & - & 0.47 & 0.50 \\
\midrule
$\downarrow$ Multiling. & \texttt{twitter-xlm} & \redtimes & {0.55} & 0.50 & \underline{0.58} & 0.56 \\
& \texttt{xlm-roberta} &  \redtimes & \underline{0.60} & 0.59 & \textbf{0.60} & \underline{0.57} \\
\rowcolor{Gainsboro!35}
& \texttt{Sentiment Projection} & \greencheck & \textbf{0.66} & \textbf{0.68} & \textbf{0.60} & \textbf{0.65}* \\
\midrule
$\downarrow$ Danish & \texttt{danish-sentiment} & \redtimes & 0.53 & 0.47 & \underline{0.58} & 0.55 \\
& \texttt{da-sentiment-base} & \redtimes & 0.23 & 0.43 & 0.08 & 0.10 \\
& \texttt{MeMo-BERT-SA} & \redtimes & 0.48 & \underline{0.67} & 0.25 & 0.24 \\
\bottomrule
\end{tabular}
\caption{Spearman correlations in the \textit{Fiction4} corpus \textit{across languages}. \underline{Columns from left to right}:  Overall evaluation on \textbf{Multilingual} dataset (English and Danish); evaluation of the \textbf{Danish set} ($n=2,800$); evaluation of the \textbf{English set} ($n=3,500$); lastly, the evaluation of \textbf{Translated} set. \underline{On top}, Inter Rater Reliability -- Spearman's $\rho$ and Krippendorff's $\alpha$. The best model performance per setting is in bold, and the follow-up is underlined. \textbf{*} There might be minimal influx in correlation caused by the concept vector being defined by untranslated sentences that are included after translation.}
\label{tab:sentiment_results_language}
\end{table*}

\autoref{tab:sentiment_results_language} demonstrates that our Sentiment Projection model leads baselines in both multilingual and Danish-only evaluations. This gain likely reflects our use of a multilingual encoder for sentence embeddings and a ``concept vector'' defined over a multilingual corpus. Concretely, Sentiment Projection attains Spearman’s $\rho=0.68$ on the Danish subset (Fairytales + Hymns) versus $\rho=0.58$ for the runner-up, and delivers a $\rho=0.06$ absolute improvement in the overall multilingual setting.

We test our model for its generalization across time periods in \autoref{tab:fiction4_corr_per_genre}, where danish hymns and fairytales represent historical language with texts from the 18-19\textsuperscript{th} century. The Sentiment Projection model shows no signs of reduced performance when processing older texts and outperforms the follow-up model by $\rho=0.12$ in the Hymns genre.

Notably, \texttt{twitter-xlm} model appears to perform slightly better on sentences translated to English than on their original Danish, as seen in \autoref{tab:sentiment_results_language}. This may indicate that Google Translate renders language in updated, contemporaneous forms, similar to the Twitter data used for model training. We see the same tendency (surprisingly) for the \texttt{danish-sentiment} model, i.e., better performance when Danish sentences were translated to English. In contrast, Sentiment Projection performs slightly better on the Danish set in its original form than when it is translated to English -- which we consider validates its capacity to process older forms reliably.

\subsection{Performance on literary and non-literary contemporary data}
 To make sure that our model does not overfit its sentiment vector to the in-context sentiment cues of the stories in the \textit{Fiction4} corpus, we tested it against the \textit{EmoBank} dataset -- which indexes contemporary literary and non-literary data. All Multilingual and dictionary-based models were tested for their correlation with the human gold standard of the \textit{EmoBank} dataset.
 The Sentiment Projection Model still achieved the highest overall correlation with human ratings. Although it shows a lower correlation for a few genres (i.a., Letters), it still appears to generalize well to contemporary out-of-training distribution data. It should be noted that the model outperforms the other models the most in the fiction genre, indicating that the sentiment vector may be slightly fine-tuned or overfit to fiction-specific sentiment indicators. While this can also be a drawback, it supports the idea that domain-specific sentiment analysis can be highly beneficial. For example, a sentiment analysis method for fiction should be sensitive to the specific sentiment cues (like omission, implicitness, concrete and object-based, etc.), rarer in other genres \cite{bizzoni_feldkamp_sea_2024}. \textcite{feldkamp2024sentiment} suggests that travel guides use similar mechanisms -- sentiment is evoked through unsentimental, descriptive, and concrete detail. The fact that Sentiment Projection performs well also for both genres suggests it captures this kind of indirect sentiment expression.

\begin{table}[ht]
    \centering
    \footnotesize
    \begin{tabular}{l|c|c|cccccc}
    \toprule
            & Scalar & Overall & Letters & Blog & Newspaper & Essays & Fiction & Travelguides \\
    \midrule
    Human IRR $\alpha$ & \greencheck & 0.34 & 0.34 & 0.31 & 0.29 & 0.31 & 0.35 & 0.23 \\
    \midrule
    \midrule
       \texttt{vader} & \greencheck & 0.43 & 0.47 & 0.41 & 0.42 & 0.32 & 0.37 & 0.35 \\
       \texttt{syuzhet} & \greencheck & 0.46 & 0.47  & 0.37 & 0.42 & 0.37 & 0.43 & 0.37 \\
        \midrule
        \texttt{twitter-xlm} & \redtimes & 0.64 & \textbf{0.69} & \textbf{0.65} & 0.61 & \textbf{0.59} & \underline{0.57} & 0.48\\
        \texttt{xlm-roberta} & \redtimes & \underline{0.65} & \underline{0.68} & \textbf{0.65} & \underline{0.65} & \underline{0.58} & 0.56 & \underline{0.49} \\
        \rowcolor{Gainsboro!35}
        \texttt{Sentiment Projection} & \greencheck & \textbf{0.67} & 0.62 & 0.61 & \textbf{0.66} & 0.53 & \textbf{0.64} & \textbf{0.52} \\
        \bottomrule
    \end{tabular}
        \caption{Spearman correlations on the \texttt{EmoBank} sentences ($n=8,870$) \textit{across domains}. On top: Inter Rater Reliability (Krippendorff's $\alpha$).}
    \label{tab:EmoBank_results}
\end{table}

\section{Discussion \& conclusions}
As seen in \autoref{tab:sentiment_results_language} and \ref{tab:EmoBank_results}, the proposed Sentiment Projection model performs on par with or better than the contemporary state-of-the-art methods. Moreover, Sentiment Projection allows for a \emph{smooth continuous output}. In contrast, methods converting model output are not continuous in practice, but rather return noncontinuous tri-modal distributions (\autoref{fig:scatterplots_corr}). While both methods correlate highly with the human golden standard, approaching the inter-rater correlation, it appears that the \emph{Sentiment Projection approach more closely resembles} the sentiment distribution of human ratings.  

Furthermore, the Sentiment Projection method can be trained on multilingual data using a multilingual language model, allowing for a \emph{language-agnostic sentiment prediction model} that also reliably handles historical variants. The Sentiment Projection was solely defined by its concept vector, based on sentences from the \textit{Fiction4} dataset, half of which were in Danish, yet it still outperforms other models.

While this paper corroborates the findings of \autocite{feldkamp_comparing_2024}, showing that translation (even without a quality check) to English increases the similarity of human and transformer-model scores, it also shows that this is not the case for Sentiment Projection, which performs slightly better on the original (Danish) sentences.

Finally, the workflow presented in \autoref{fig:projection_workflow} has been used to design a sentiment model, but allows easy generalization to other concepts of choice. The method could also work for other emotional concepts, such as emotion recognition, language detection, or abstract concepts like a nature-to-industry gradient. We encourage curious readers to search for inspiration for potential vectors in \textit{Linear Representation Hypothesis} \autocite{Peterson-2020-Parallelograms-Revisited} and \textit{Steering Vector} \autocite{wehner_taxonomy_2025} literature. Due to the flexible nature of the algorithm, there is no rigid lower boundary on the number of training points required for a stable vector, although the chances of over-representing the non-concept context of training sentences naturally increase as the number of sentences decreases. A future empirical investigation of the stability of the vector when using smaller training sets would be useful.  


\section*{Acknowledgements}

\printbibliography

\appendix

\clearpage
\section{Models} 
\label{appdx:A}

\begin{table}[ht]
    \footnotesize
    \centering
    \begin{tabular}{llr}
    \textbf{Type} & & \textbf{Shorthand, Modelname \& URLs} \\
    \toprule
    $\downarrow$ Encoder & &  \\
       & Shorthand & Sentiment Projection \\
       & Name & \texttt{Sentiment Projection using paraphrase-multilingual-mpnet-base-v2} \\
       &  URL & \textcolor{blue}{https://huggingface.co/sentence-transformers/paraphrase-multilingual-mpnet-base-v2} \\
        \midrule
        \midrule
      $\downarrow$ Multiling. &  & \\
      & Shorthand & twitter-xlm \\
      & Name & \texttt{cardiffnlp/twitter-xlm-roberta-base-sentiment-multilingual} \\ 
      & URL & \textcolor{blue}{https://huggingface.co/cardiffnlp/twitter-xlm-roberta-base-sentiment-multilingual} \\
      \midrule
      & Shorthand & xlm-roberta \\
     & Name &   \texttt{cardiffnlp/xlm-roberta-base-sentiment-multilingual} \\
     & URL &   \textcolor{blue}{https://huggingface.co/cardiffnlp/xlm-roberta-base-sentiment-multilingual} \\
        \midrule
        \midrule
    $\downarrow$ Danish & & \\
     & Shorthand & danish-sentiment \\
     & Name & \texttt{vesteinn/danish\_sentiment} \\
     & URL & \textcolor{blue}{https://huggingface.co/vesteinn/danish\_sentiment} \\
         \midrule
     & Shorthand & da-sentiment-base \\
     & Name & \texttt{alexandrainst/da-sentiment-base} \\
     & URL & \textcolor{blue}{https://huggingface.co/alexandrainst/da-sentiment-base} \\
         \midrule
     & Shorthand & MeMo-BERT-SA \\
     & Name & \texttt{MiMe-MeMo/MeMo-BERT-SA} \\
     & URL & \textcolor{blue}{https://huggingface.co/MiMe-MeMo/MeMo-BERT-SA} \\
        \bottomrule
    \end{tabular}
    \caption{Full model names \& details.}
    \label{tab:modelnames}
\end{table}

\clearpage
\section{Score distribution}

\begin{figure}
    \centering
    \includegraphics[width=1\linewidth]{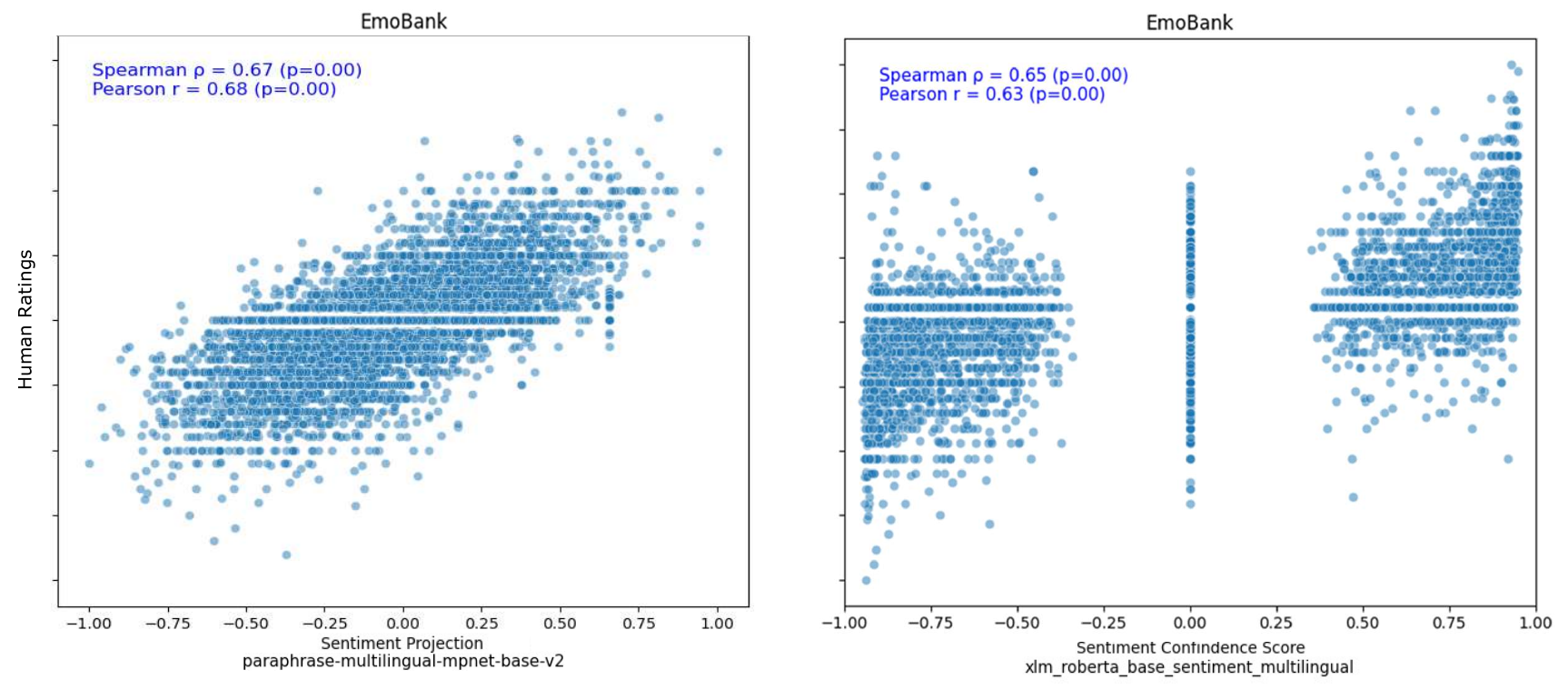}
    \caption{Scatterplot of Sentiment Projection \texttt{xlm-roberta} for \textit{EmoBank} Data.}
    \label{fig:EmoBank_Scatter}
\end{figure}

\begin{figure}
    \centering
    \includegraphics[width=0.9\linewidth]{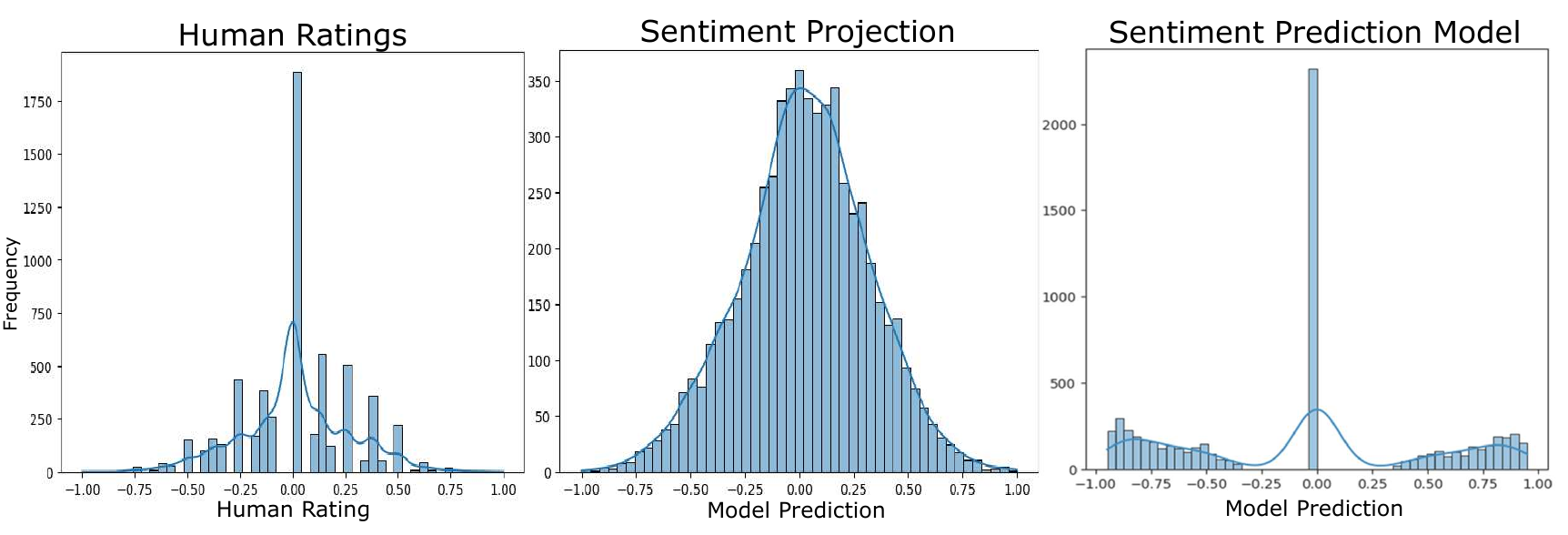}
    \caption{Histograms of respectively Human raters, sentiment projection model and \texttt{xlm-roberta}'s predictions for the \textit{Fiction4} test-set. This plot should be interpreted in conjunction with \autoref{fig:scatterplots_corr} and \autoref{fig:EmoBank_Scatter}. It visualizes that the \texttt{xlm-roberta} model follows the human trend of predicting completely neutral sentences. The Sentiment Projection predicts mostly neutral sentences, as hoped, but follows a bell-curve that becomes visible in human ratings, as the number of raters increases, see \autoref{fig:EmoBank_Hist}.}
    \label{fig:Prediction_Histograms}
\end{figure}

\begin{figure}
    \centering
    \includegraphics[width=0.7\linewidth]{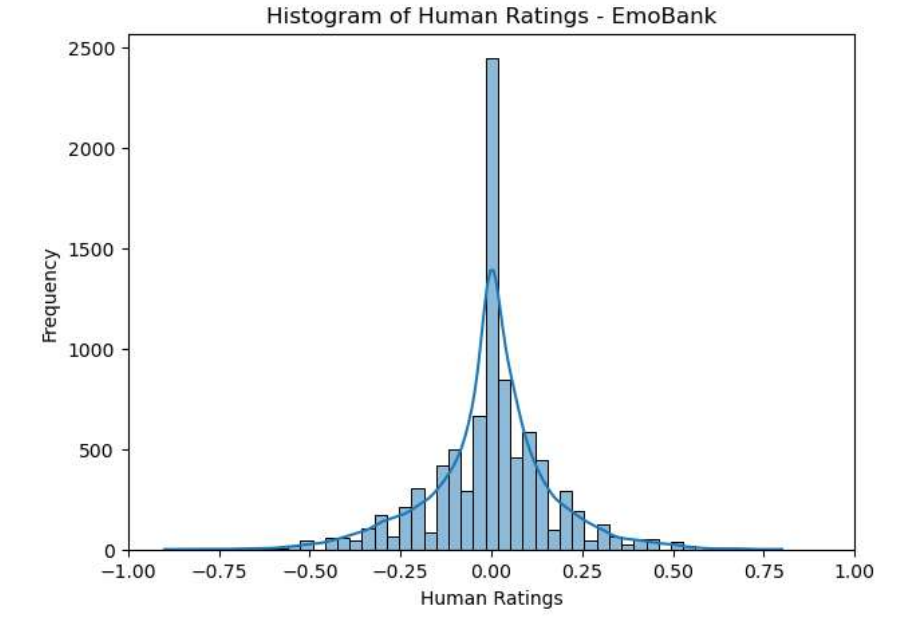}
    \caption{Histogram of human ratings. As ratings become the average of 10 raters, it approaches a more continuous bell-shaped form, in comparison to the 3-rater average depicted in the \textit{Human Rating} plot in \autoref{fig:Prediction_Histograms}.}
    \label{fig:EmoBank_Hist}
\end{figure}

\clearpage
\section{Algorithm}
\label{sec:algorithm}

The following algorithm formally describes the procedure for defining and applying a concept vector by using labeled sentence embeddings. 

\begin{algorithm}
  \caption{Concept Vector Projection}\label{conceptvector}
  \begin{algorithmic}[0]
  \State \textbf{Input:} \\
  $\mathcal{M}$ = Language Model \\
  $\mathcal{S}$ = A set of categorically labeled sentences  $s_i \in \{\text{positive}^+, \text{negative}^-, \text{neutral}^\varnothing, \text{unknown}^?\}$
  \State \textbf{Output:} \\
  $\hat{\mathbf{v}}$ = Concept vector  \\
  score($s_i$) = projection scores for unknown sentences
 \State \textbf{Computation:}
 \end{algorithmic}
  \begin{algorithmic}[1]
  \State Embed all sentences: $\mathbf{e}_i = \mathcal{M}(s_i)$
  \State $P^+ \gets \{\mathbf{e}_i \mid s_i = \text{positive}\}$
  \State $N^- \gets \{\mathbf{e}_i \mid s_i = \text{negative}\}$
  \State Compute means: $\vec{\mu_{s^{+}}} = \text{mean}(P^+)$, $\vec{\mu_{s^{-}}} = \text{mean}(N^-)$
  \State Compute concept vector: $\vec{\mathbf{v}} = \vec{\mu_{s^{+}}} - \vec{\mu_{s^{-}}}$
  \State Normalize: $\hat{\mathbf{v}} = \frac{\vec{\mathbf{v}}}{\|\vec{\mathbf{v}}\|}$
  \For{\textbf{each} embedding $\mathbf{e}_i$} 
    \State $\text{score}(s_i) = \mathbf{e}_i \cdot \hat{\mathbf{v}}$   \hfill // Embedding projection
  \EndFor
  \end{algorithmic}
\end{algorithm}
\end{document}